\documentclass[11pt]{article}

\usepackage[final]{acl}

\usepackage{times}
\usepackage{latexsym}
\usepackage{enumitem}

\usepackage[T1]{fontenc}

\usepackage[utf8]{inputenc}

\usepackage{microtype}

\usepackage{inconsolata}

\usepackage{amsmath}
\usepackage{graphicx}
\usepackage{subcaption}
\usepackage{booktabs}

\title{The Model's Language Matters: A Comparative Privacy Analysis of LLMs}

\author{
    Abhishek K. Mishra \and
    Antoine Boutet \and
    Lucas Magnana \\
    Inria, INSA Lyon, CITI \\
    \texttt{\{abhishek.mishra, antoine.boutet, lucas.magnana\}@inria.fr}
}

\begin{document}
\maketitle

\begin{abstract}

Large Language Models (LLMs) are increasingly deployed across multilingual applications that handle sensitive data, yet their scale and linguistic variability introduce major privacy risks. Mostly evaluated for English, this paper investigates how language structure affects privacy leakage in LLMs trained on English, Spanish, French, and Italian medical corpora. We quantify six linguistic indicators and evaluate three attack vectors: extraction, counterfactual memorization, and membership inference. Results show that privacy vulnerability scales with linguistic redundancy and tokenization granularity: \texttt{Italian} exhibits the strongest leakage, while \texttt{English} shows higher membership separability. In contrast, \texttt{French} and \texttt{Spanish} display greater resilience due to higher morphological complexity. Overall, our findings provide the first quantitative evidence that \textit{language matters} in privacy leakage, underscoring the need for language-aware privacy-preserving mechanisms in LLM deployments. 
\end{abstract}

\section{Introduction}

Rapid advances in natural language processing (NLP) have fueled its adoption in many industries worldwide. Large language models (LLMs) such as BERT and GPT have been pre-trained at great expense on countless unlabeled datasets extracted from the internet.
While these models represent incredible potential and promises, their large-scale deployment and their complexity, as well as the fact that they interact with and potentially influence individuals, raise multiple security and privacy concerns~\cite{das2025security}.

The attack surface on models is still poorly understood \cite{10.1145/3531146.3533088,lehman-etal-2021-bert,duprieu:hal-04782667,chen-etal-2024-learnable}.
A number of threats are related to the memorization and possible leakage of sensitive information used during model training, such as data reconstruction and membership inference (i.e., identifying elements used during the training or the fine-tuning).
Memorization of information by a model is not a problem in itself. 
However, this memorization becomes a problem when the training information is not generalized enough by the model which reproduces large portions of training data verbatim or discloses some sensitive information~\cite{kassem-etal-2023-preserving, wu-etal-2023-depn}.

Most privacy risk assessment work has been conducted on English texts~\cite{li2024llm, shanmugarasa2025sok}.
However, the language of the texts, their structures, and their characteristics inherently impact LLM memorization and, consequently, privacy risks.
Although the language considered is well known to potentially introduce bias in some results, the impact of language on privacy risks has not yet been explored to our knowledge.
To overcome this limitation, in this paper, we empirically explore the impact of language on privacy risks associated with LLMs.
We also analyze the main characteristics and structures of each language and link them to various privacy vulnerabilities.
Specifically, we comprehensively assess the privacy of LLMs fine-tuned on English, Spanish, French, and Italian medical corpora using an extraction attack, a membership inference attack and counterfactual memorization.
Results show that privacy vulnerability scales with linguistic redundancy and tokenization granularity: Italian presents the highest leakage, while English has higher membership separability. In contrast, French and Spanish show greater resilience due to greater morphological complexity.

Overall, our results provide the first quantitative evidence that language is a significant factor in LLM privacy leakage.
This highlights the need to consider this factor in LLM deployment and the design or configuration of privacy-preserving mechanisms.

\section{Background and Related Work}


Large language models (LLMs) are trained on very large datasets.
For example, training chatGPT required years of crawling the Internet.
Therefore, a lot of personal data such as people’s addresses was used during training.
BERT models, on the other hand, are typically fine-tuned for specific tasks with domain-oriented data.
In the medical domain, datasets typically include sensitive patient records.
In both cases, the problem is that the models can regurgitate and leak information from the training data after deployment~\cite{extract,cooper2023report1stworkshopgenerative}.

A central question in this context concerns the extent to which language models memorize their training data~\cite{carlini2019secretsharerevaluatingtesting,extract,nasr2023scalableextractiontrainingdata,counterfactual,schwarzschild2024rethinkingllmmemorizationlens}.
However, defining memorization for language models is challenging, and many existing definitions and notions have been proposed depending on whether the memorization concerns copyrighted content or personal and sensitive content.
In relation to privacy, we can notably cite extractable memorization (Section~\ref{sec:extr}) and membership inference (Section~\ref{sec:miah}).

\subsection{\textbf{Extractable memorization}}
\label{sec:extr}


Extractable memorization is a type of attack that aims to use the model to infer information from the original data~\cite{privacy-survey}. This attack mainly concerns text generation models, such as GPT. These models are trained to produce text based on what they have seen during training. However, the model is not expected to be a basic parrot and repeat exactly the sentences it has seen. This is especially concerning if the data it is repeating is sensitive. This has been shown to be the case with GPT-2 for example, from which the names and addresses of individuals can be extracted~\cite{gpt2}.
In~\cite{extract}, the term $k-extractability$ is used to refer to the sequences that can be extracted from the model when an input sequence of length $k$ is requested. The lower the $k$, the easier it is to extract the sequence. We therefore expect a model to have the highest possible $k$ on private queries.
This measure, however, does not capture regurgitations that are not perfect, which can lead to an illusion of no extractable memory.
Compressible memorization~\cite{schwarzschild2024rethinkingllmmemorizationlens} extends this definition by evaluating how short the minimal requested sentence (or prompt) that elicits the sequence.\\

\subsection{\textbf{Membership inference}}
\label{sec:miah}



Membership inference attack~\cite{carlini2022membership,shokri2017membership,jagannatha2021membership,mireshghallah2022quantifying,10020711,hayes2025strongmembershipinferenceattacks} (MIA) is a more common inference attack in Machine Learning (ML), which aims to infer whether a specific data was used in the training data of a target model. 
%
%
There are different techniques that can be used to perform a MIA attack depending on if the adversary has an access to the model parameters (i.e., white-box access), or access to a ground-truth
subset of member and non-member samples. One technique consists to analyze the loss of member and non member samples~\cite{8429311}, another one is to use multiple shadow models~\cite{shokri2017membership,10.1145/3548606.3560675} trained to mimic the behavior of the target model on an auxiliary dataset.
An adversarial model is then trained to infer membership from the loss or from shadow models.

Another method~\cite{counterfactual} is based on comparing the performance of the target model trained on a dataset with a specific input, with a second model trained without it.
As ML models are supposed to learn general information, one piece of data (even rare, outlier or mislabeled samples) is not supposed to be memorized and significantly changed the model's performance. By repeating this operation many times with different subset, it is possible to identify counterfactually memorized data.

\section{Comparative Privacy Analysis}
\label{sec:privacyleakage}

We perform a comparative privacy analysis across four languages: \texttt{English}, \texttt{Spanish}, \texttt{French}, and \texttt{Italian}, encompassing three complementary threat models:
(i)~\textit{prompt-based extraction}, where we probe direct content leakage from generative models;
(ii)~\textit{counterfactual memorization}, where we quantify how strongly individual texts are overfit by fine-tuned models; and
(iii)~\textit{membership inference}, where we test whether a model exposes the presence of individual samples in its training set.
Together, these analyses provide a unified view of surface-level and latent memorization behaviour across languages and architectures. 


\subsection{Experimental Setup}
\label{sec:methodology} 

\paragraph{Datasets.}
We employ a corpus to capture both controlled and large-scale multilingual behavior. The \texttt{HiTZ Multilingual Medical Corpus}~\footnote{\begin{tiny} \url{https://huggingface.co/datasets/HiTZ/Multilingual-Medical-Corpus}\end{tiny}} provides over $3$ million translated medical sentences in \texttt{English}, \texttt{Spanish}, \texttt{French}, and \texttt{Italian}. We select 10k sentences from the corpus in this analysis, as it is large enough for privacy assessment while accounting for the limited computational resources that we have. 



\paragraph{Model Selection and Training.}
We evaluate both encoder-only (\texttt{BERT}-style) and decoder-only (\texttt{GPT}-style) architectures to contrast their privacy behaviors across tasks. 
Encoder models are assessed through classification-based membership inference and counterfactual memorization, 
while decoder models are probed via generative extraction, providing a complementary view of implicit versus explicit memorization dynamics. 

For encoder-only architectures, we fine-tune one pre-trained model per language on a medical classification task: 
\texttt{bert-base-uncased}\footnote{\begin{tiny}
    \url{https://huggingface.co/bert-base-uncased}\end{tiny}} (English), 
\texttt{dccuchile/bert-base-spanish-wwm-cased}\footnote{\begin{tiny}\url{https://huggingface.co/dccuchile/bert-base-spanish-wwm-cased}\end{tiny}} (Spanish), 
\texttt{almanach/camembert-base}\footnote{\begin{tiny}\url{https://huggingface.co/almanach/camembert-base}\end{tiny}} (French), 
and \texttt{Musixmatch/umberto-commoncrawl-cased-v1}\footnote{\begin{tiny}\url{https://huggingface.co/Musixmatch/umberto-commoncrawl-cased-v1}\end{tiny}} (Italian). 
For decoder-only architectures, used in extraction attacks, we fine-tune \texttt{distilgpt2}\footnote{\begin{tiny}\url{https://huggingface.co/distilgpt2}\end{tiny}} (English), 
\texttt{DeepESP/gpt2-spanish}\footnote{\begin{tiny}\url{https://huggingface.co/DeepESP/gpt2-spanish}\end{tiny}} (Spanish), 
\texttt{dbddv01/gpt2-french-small}\footnote{\begin{tiny}\url{https://huggingface.co/dbddv01/gpt2-french-small}\end{tiny}} (French), 
and \texttt{LorenzoDeMattei/GePpeTto}\footnote{\begin{tiny}\url{https://huggingface.co/LorenzoDeMattei/GePpeTto}\end{tiny}} (Italian). 
All models are trained using identical hyperparameters (batch size, learning rate, and number of epochs) across languages to ensure comparability. 
Each dataset is randomly split into $80\%$ for training and $20\%$ for testing, maintaining consistent data exposure across experiments. 

\paragraph{Attack Setup.}
\begin{itemize}[leftmargin=*]
    \item \textbf{Extraction attacks: } We perform prompt-based extraction attacks to evaluate explicit surface leakage in generative models.
    Our approach conditions a fine-tuned decoder model on partial text fragments
    and measures how often it regenerates exact or near-exact spans from the training corpus.
    We systematically vary the prompt fraction in $\{5,12, 25, 37\}$ to examine how prompt length
    influences extraction behaviour.
    Unlike prior optimization-based extraction methods, our strategy requires no gradient access
    and scales efficiently across multiple languages.
    We additionally quantify the number and diversity of unique extractions as a function of prompt size,
    providing a direct signal of language-dependent memorization risk. 

    \item \textbf{Counterfactual memorization: } 
    We quantify instance-level overfitting by computing a \textit{counterfactual memorization score} for each document in the HiTZ Multilingual Medical Corpus.
    Each model is fine-tuned on a 9-class \emph{length-binned text classification} task, where labels correspond to decile-based token length bins.
    For each text, the counterfactual score is defined as the difference between the mean sigmoid loss of models that \emph{saw} the text during training and those that did not.
    This metric extends standard memorization analysis by capturing the intensity of instance-level overfitting.
    We train an ensemble of ten independently seeded sequence classifiers per language, based on BERT-family encoders, to ensure stable counterfactual estimates.
    The 95th percentile of the resulting score distribution is used to flag highly memorized instances.
    We further compute empirical CDFs over surface-level statistics (e.g., sentence length, word count, unique words) to relate memorization strength to linguistic and morphological characteristics (Table~\ref{tab:hitz_metrics}).  

    \item \textbf{Membership inference: } 
    We evaluate membership inference on the same fine-tuned classification models, using shadow models trained to replicate the target model’s learning dynamics.
    Attackers exploit differences in prediction confidence distributions to distinguish ``in-training’’ versus ``out-of-training’’ samples.
    This setup targets encoder-only architectures and quantifies privacy leakage arising from confidence calibration and representation separability.
    Since the underlying classification task is language-agnostic (based on text length bins), it provides a controlled baseline for assessing how linguistic structure influences susceptibility to membership inference. 
\end{itemize}



\subsection{Extraction Attack}
\label{sec:extraction}

\begin{figure}[t!]
    \centering
    \includegraphics[width=0.5\textwidth]{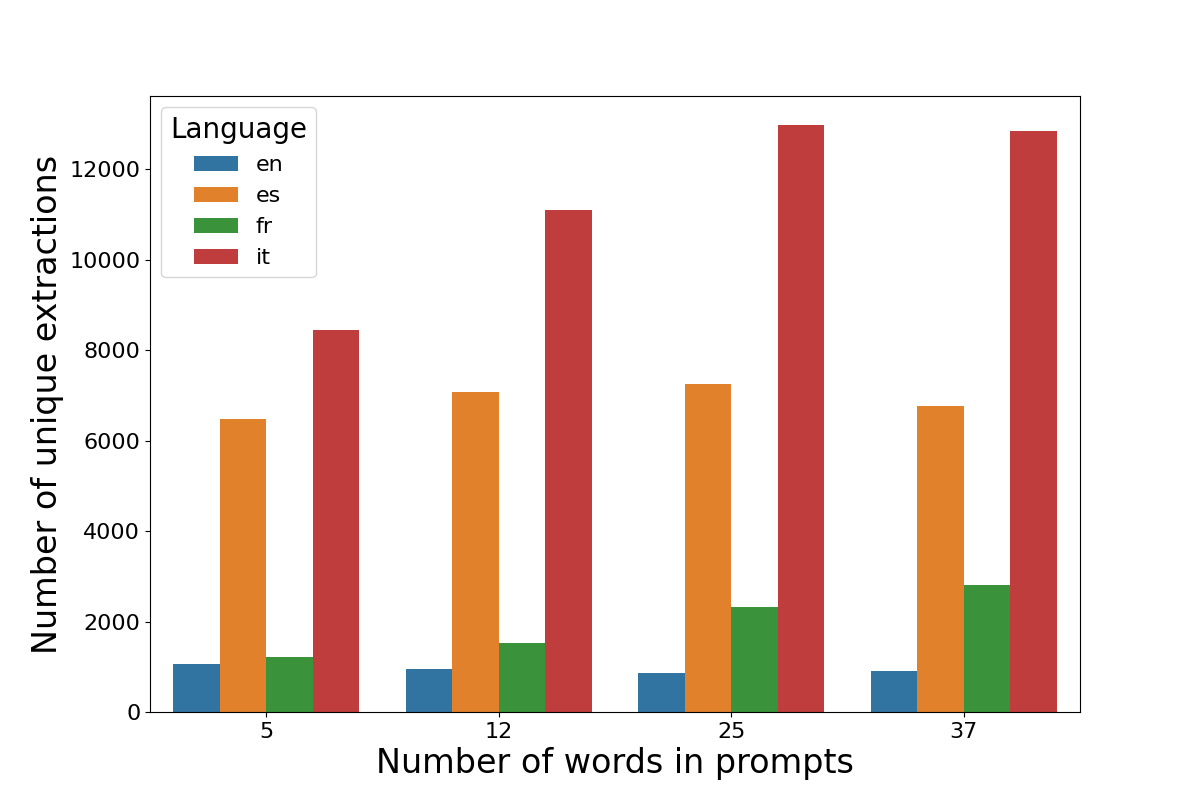}
    \caption{Number of unique extractions across languages and prompt sizes: longer prompts increase extraction risk in general.}
    \label{fig:barplot-multi}
\end{figure}

\begin{figure}[t!]
    \centering
    \includegraphics[width=0.48\textwidth]{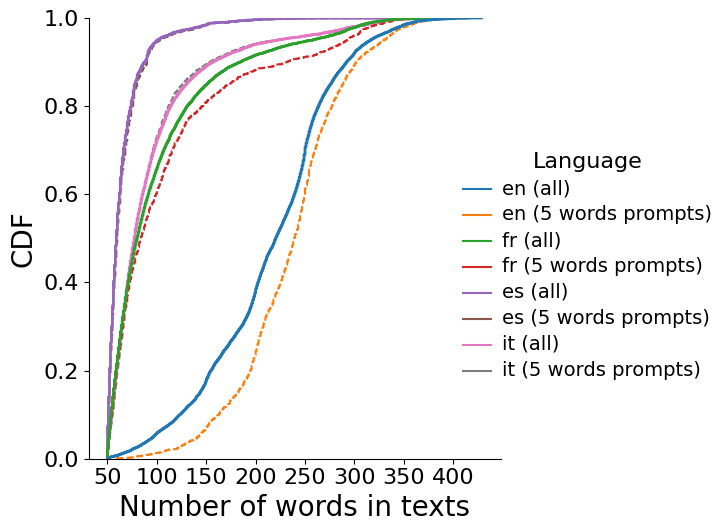}
    \caption{Cumulative distribution of text lengths for all versus extracted samples.}
    \label{fig:cdf-match-len}
\end{figure}

We probe surface-level memorization through prompt-conditioned extraction attacks, where partial context is provided to a generative model to elicit verbatim continuations. 
Figure~\ref{fig:barplot-multi} quantifies the number of \emph{unique} extractions across languages and prompt sizes, while Figure~\ref{fig:cdf-match-len} reports the cumulative distribution of text lengths for all sentences versus those appearing among extracted samples (with a short 5-word prompt).

We observe marked cross-linguistic differences. 
At minimal prompts (i.e., 5 words), \texttt{English} produces fewer than 1,000 unique extractions, suggesting relatively low surface-level leakage under constrained context. 
In contrast, \texttt{Spanish} already yields over 6,000 unique spans, and \texttt{Italian} surpasses 8,000, indicating greater sensitivity to minimal cues. 
As prompt size increases to 12 and 25 words, Italian extractions rise sharply, peaking at over 13,000 unique spans, while Spanish stabilizes around 7,000. 
\texttt{French}, by comparison, remains substantially lower throughout, increasing from roughly 1,200 to 2,700 extractions. 
These patterns reveal that certain languages (ES, IT) sustain or amplify leakage as prompts grow, whereas English shows an early saturation and subsequent decline in extraction counts with larger context windows.

Moreover, further analysis reveals that longer texts are more prone to extraction even under short prompts. 
As illustrated in Figure~\ref{fig:cdf-match-len}, the CDFs for the extracted texts (i.e., using 5-word prompts) closely follow or are slightly shifted to the right of the overall corpus distributions, indicating that the extracted samples tend to contain more words on average. 
This demonstrates that extraction behavior with short prompts is not biased by sentence length: even minimal context captures the same cross-linguistic tendencies observed in Figure~\ref{fig:barplot-multi}. 
Consequently, the higher number of extractions in \texttt{Spanish} and \texttt{Italian} cannot be attributed to prompt selection, but rather reflects their intrinsic linguistic and structural susceptibility to memorization.

\begin{figure}[t!]
    \centering
    \includegraphics[width=0.4\textwidth]{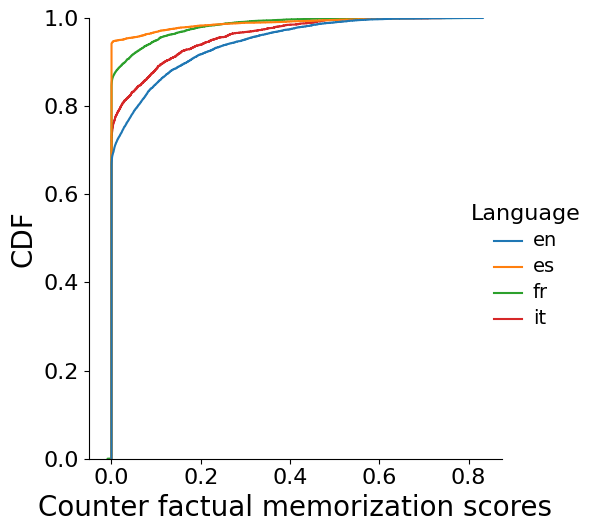}
    \caption{Distribution of counterfactual memorization scores across languages. Most points lie near zero; EN and IT display extended positive tails, FR shows rare high outliers, and ES remains the most compact.}
    \label{fig:memorization-multi}
\end{figure}

\begin{figure*}[t!]
    \centering
    \subfloat[\texttt{English}]{\includegraphics[width=0.24\textwidth]{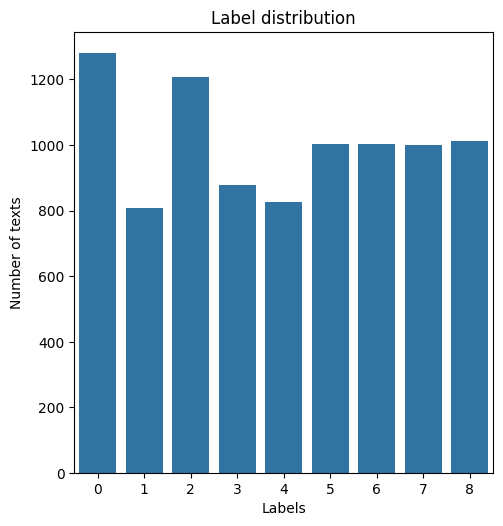}}
    \subfloat[\texttt{Spanish}]{\includegraphics[width=0.24\textwidth]{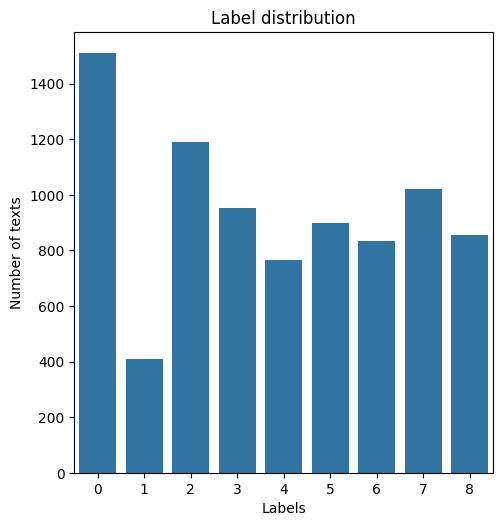}}
    \subfloat[\texttt{French}]{\includegraphics[width=0.24\textwidth]{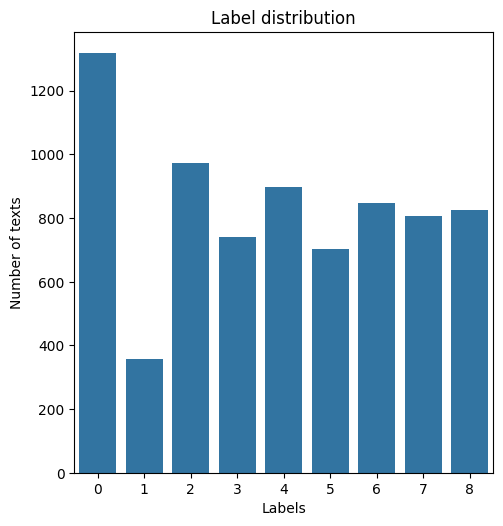}}
    \subfloat[\texttt{Italian}]{\includegraphics[width=0.24\textwidth]{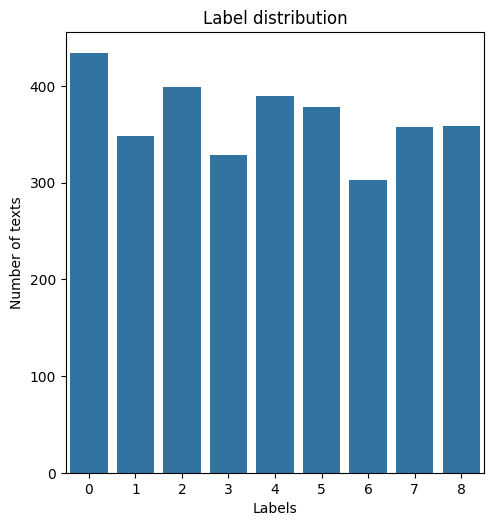}}
    \caption{Label distributions used for memorization scoring: balanced bins across languages confirm that score variations are not due to class imbalance.}
    \label{fig:label-dist}
\end{figure*} 

\begin{figure*}[t!]
    \centering
    \subfloat[\texttt{English}]{\includegraphics[width=0.24\textwidth]{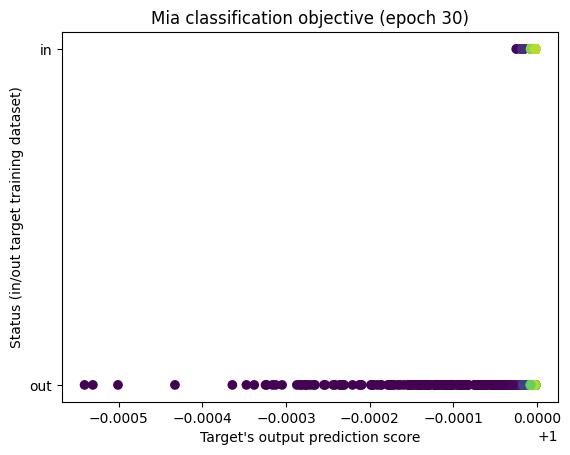}}
    \subfloat[\texttt{Spanish}]{\includegraphics[width=0.24\textwidth]{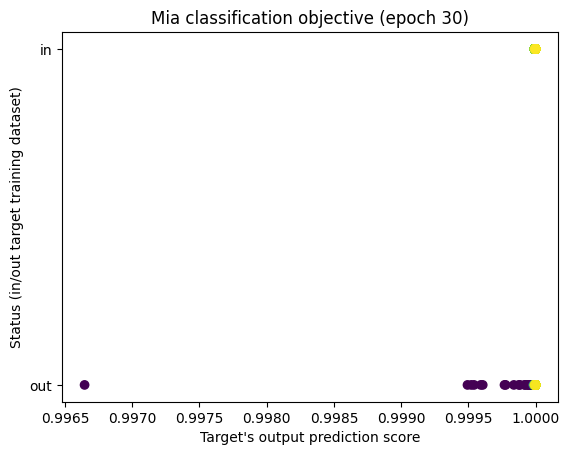}}
    \subfloat[\texttt{French}]{\includegraphics[width=0.24\textwidth]{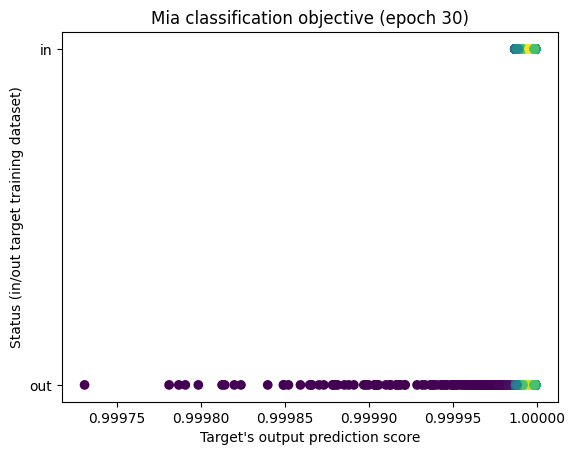}}
    \subfloat[\texttt{Italian}]{\includegraphics[width=0.24\textwidth]{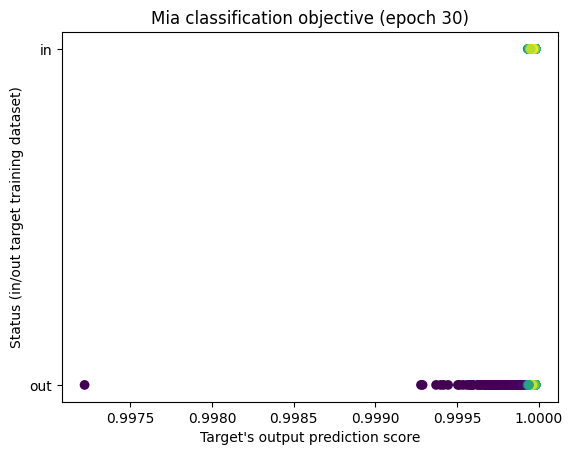}}
    \caption{Separability of ``in'' vs.\ ``out'' samples at epoch~30 under MIAs: larger gaps indicate higher risk. \texttt{English} exhibits the most distinct separation between training and test data, while \texttt{French} shows the greatest overlap, indicating stronger generalization.}
    \label{fig:mia-multi}
\end{figure*}

\subsection{Counterfactual Memorization}
\label{sec:memorization}

The counterfactual memorization score measures the change in loss between models that \emph{saw} a text during training and those that did not. 
This metric captures how strongly each instance is memorized relative to a counterfactual baseline. 
Figure~\ref{fig:memorization-multi} reports the score distributions across languages, while Figure~\ref{fig:label-dist} confirms that label distributions are balanced and therefore do not confound memorization effects.

The results reveal that most samples cluster around zero in all languages, indicating that the majority of instances are \textbf{not explicitly memorized}. 
However, language-specific deviations appear in the positive tail of the distribution. 
\texttt{Spanish} exhibits the narrowest spread, with over 95\% of samples scoring below $0.02$, suggesting minimal overfitting and strong generalization. 
\texttt{English} and \texttt{Italian}, by contrast, show moderate positive tails extending up to $0.08$--$0.10$, indicating that 5--8\% of samples exhibit measurable memorization. 
Finally, \texttt{French} displays a distinctive pattern: while its median score remains low, it contains rare but pronounced outliers that exceed $0.15$, pointing to isolated cases of high-confidence recall. 

We further analyze the label distributions used in the counterfactual memorization experiments to verify dataset balance across languages (Figure~\ref{fig:label-dist}). 
Although minor variations exist, all languages maintain a roughly uniform spread over the nine label bins, ensuring that observed memorization trends are not artifacts of label skew.

Quantitatively, \texttt{English} exhibits a mildly-skewed distribution. 
\texttt{Spanish} shows a similar pattern but with a more pronounced peak at label~0 ($\sim$1{,}500 texts) and a small dip around label~1 ($\sim$400 texts). 
\texttt{French} follows a nearly identical trend, with its most frequent label~0 ($\sim$1{,}300$+$ texts) and the least represented label~1 ($\sim$350$-$400 texts). 
In contrast, \texttt{Italian} displays the most balanced profile, with all labels ranging between 300 and 450 samples and no extreme outliers.

These distributions confirm that the memorization differences reported in Figure~\ref{fig:memorization-multi} cannot be attributed to unbalanced label frequencies. 
While \texttt{English}, \texttt{Spanish}, and \texttt{French} exhibit mild concentration toward lower labels, all maintain sufficient coverage of the label space to ensure unbiased counterfactual comparisons. 
The flat histogram of the \texttt{Italian} dataset further demonstrates that even with a highly uniform label representation, moderate memorization persists, reinforcing that linguistic and structural factors, rather than label imbalance, drive the cross-lingual variability observed in memorization strength.

\subsection{Membership Inference Attack}
\label{sec:mia}

We evaluate the susceptibility of our models to membership inference attacks (MIAs) by analyzing whether an adversary can distinguish samples that were part of the training set (\textit{in}) from those that were not (\textit{out}). 
Our analysis focuses on encoder-based models (\texttt{BERT}-family) fine-tuned for classification in each language.

To simulate a realistic adversary, we train a shadow model following the same architecture and optimization procedure as the target model but using a controlled dataset composed of both training (\textit{in}) and test (\textit{out}) samples. 
The attacker then observes the per-sample confidence scores produced by the shadow and target models to learn a decision boundary distinguishing ``in'' from ``out'' samples. 
This boundary is learned using an \texttt{XGBoost} classifier trained on confidence distributions across epochs (1–30), as the separability between \textit{in} and \textit{out} typically increases with training progression.

\paragraph{Training dynamics.}
As expected, we observe that model confidence for training data progressively diverges from that of unseen data as training advances. 
Early in training (epochs 1–5), the overlap between \textit{in} and \textit{out} confidence distributions remains substantial, making inference difficult. 
By epoch~30, however, clearer separation emerges, with train samples forming high-confidence clusters and test samples occupying lower ranges. 
This evolution highlights how overfitting amplifies membership signal leakage over time.

\paragraph{Cross-lingual separability.}
Figure~\ref{fig:mia-multi} visualizes the final in/out confidence distributions at convergence. 
\texttt{English} exhibits the most distinct separation, where the attacker achieves an MIA accuracy of $0.59$, with train-set precision of $0.54$ and test-set precision of $0.98$. 
This indicates considerable memorization effects and high confidence calibration differences between seen and unseen samples. 
\texttt{Spanish} and \texttt{Italian} occupy an intermediate regime, achieving accuracies around $0.51$--$0.54$, where partially overlapping distributions still expose mild but detectable membership traces. 
\texttt{French} demonstrates the tightest overlap between distributions, yielding the lowest attack accuracy ($0.50$), suggesting better generalization and minimal membership signal leakage.

\paragraph{Implications.}
These findings highlight a positive coupling between overfitting and membership vulnerability: models that exhibit pronounced memorization behavior (e.g., \texttt{English}, \texttt{Italian}) are also the most susceptible to membership inference. Languages like \texttt{French}, which generalize more smoothly, naturally mitigate this exposure.

\section{Privacy Implications of the Language Structures}
\label{language-structures}


\subsection{Language Characteristics}
\label{sec:lang-characteristics}

To capture linguistic properties that may affect memorization and extraction, we compute six structural and morphological indicators. Each metric highlights a specific typological feature that could modulate privacy leakage in LLMs.

\paragraph{Morphological complexity.}
We measure the average number of inflectional variants per lemma, reflecting how flexional morphology increases linguistic variability~\citep{juola1998measuring,brown2018simple,marzi2020inflection,ccoltekin2023complexity}:
\begin{equation}
\footnotesize
\mathcal{M} = \frac{1}{|V|} \sum_{w \in V} |\mathcal{I}(w)|,
\end{equation}
where $V$ is the lemma vocabulary and $\mathcal{I}(w)$ denotes the set of inflected forms of lemma $w$. 

\paragraph{Syntactic entropy.}
This measures word-order variability and structural diversity in dependency relations~\citep{futrell2019syntactic,marcolli2016syntactic,levshina2019token}:
\begin{equation}
\footnotesize
\mathcal{S} = - \sum_{r \in R} P(r)\log P(r),
\end{equation}
where $R$ is the set of syntactic relations and $P(r)$ their empirical probabilities.

\paragraph{Redundancy and predictability.}
We quantify local contextual predictability through mutual information between neighboring tokens~\citep{wolf2023quantifying,li1989mutual}:
\begin{equation}
\footnotesize
\mathcal{R} = \frac{1}{N}\sum_{i=1}^{N} I(w_i; w_{i-1}, w_{i+1}),
\end{equation}
where $I$ denotes mutual information and $N$ the number of tokens. Higher $\mathcal{R}$ implies greater repetition and potential for memorization.

\paragraph{Tokenization characteristics.}
The average word length serves as a proxy for token fragmentation and morphological density~\citep{tamang2024evaluating}:
\begin{equation}
\footnotesize
\mathcal{T} = \frac{1}{|W|}\sum_{w \in W} \text{len}(w),
\end{equation}
where $\text{len}(w)$ represents the character length of word $w$.

\paragraph{Capitalization and orthography.}
We estimate the proportion of capitalized words, which often correspond to named entities and thus correlate with identifiable content~\citep{beaufays2013language}:
\begin{equation}
\footnotesize
\mathcal{C} = \frac{1}{|W|}\sum_{w \in W} \mathbf{1}[\text{isCapitalized}(w)].
\end{equation}

\paragraph{Vocabulary richness.}
Lexical diversity is represented by the type–token ratio, reflecting the productivity and variability of vocabulary~\citep{lu2012relationship}:
\begin{equation}
\footnotesize
\mathcal{D} = \frac{|V|}{|W|},
\end{equation}
where $|V|$ is the number of unique word types and $|W|$ the total number of tokens.

These indicators collectively reveal the typological contrasts that underlie variations in memorization and extraction behaviors across languages. 
Languages characterized by higher morphological complexity and more flexible syntax tend to exhibit distinct privacy-leakage patterns compared to more analytically structured ones. 

Specifically, redundancy in linguistic structure amplifies memorization risk by reinforcing repeated patterns; 
a high capitalization rate signals greater exposure to named entities such as persons or locations, heightening the risk of sensitive data leakage; 
rich vocabulary and morphological variability may introduce natural obfuscation but simultaneously complicate de-identification; 
and finally, elevated syntactic entropy reflects greater structural diversity, increasing the likelihood of memorizing unique linguistic sequences.

\subsection{Comparing Language Characteristics}
\label{sec:compare-languages}

We compare linguistic metrics across English, Spanish, French, and Italian medical corpora from the \texttt{HiTZ} dataset to assess how language structure influences privacy leakage during LLM training. Table~\ref{tab:hitz_metrics} presents six key linguistic indicators used in this comparison.

\begin{table*}[t!]
\caption{Linguistic metrics across four languages in the HiTZ multilingual medical corpus.}
\label{tab:hitz_metrics}
\centering
\footnotesize
\begin{tabular}{lrrrrrr}
\toprule
 & \textbf{Morph. Comp.} & \textbf{Synt. Ent.} & \textbf{Redundancy} & \textbf{Avg. Word Len.} & \textbf{Cap. Rate} & \textbf{Vocab. Rich.} \\
\midrule
\texttt{English} & 1.2227 & 2.9025 & 7.7728 & 5.7750 & 0.1446 & 0.1148 \\
\texttt{Spanish} & 1.2257 & 2.8119 & 7.3764 & 5.7939 & 0.0977 & 0.1269 \\
\texttt{French}  & 1.3454 & 2.8632 & 7.2191 & 5.4756 & 0.0733 & 0.0776 \\
\texttt{Italian} & 1.1559 & 2.7822 & 8.6942 & 5.9922 & 0.1538 & 0.2193 \\
\bottomrule
\end{tabular}
\end{table*}

\begin{figure}[t!]
  \centering
  \begin{subfigure}{0.48\linewidth}
    \includegraphics[width=\linewidth]{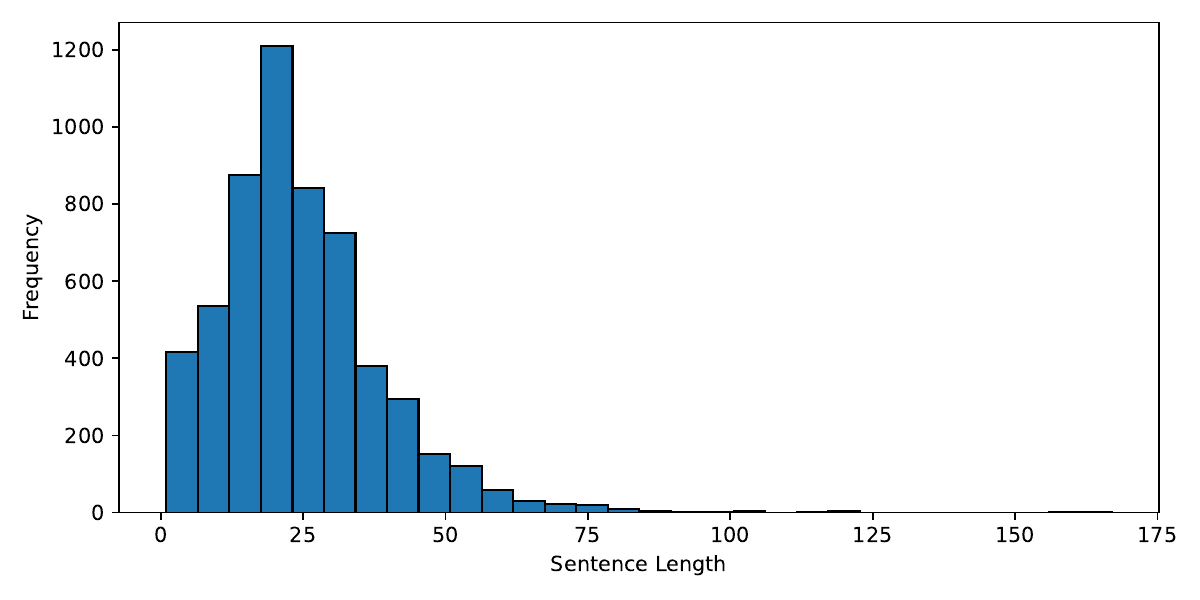}
    \caption{\texttt{English)}}
  \end{subfigure}
  \hfill
  \begin{subfigure}{0.48\linewidth}
    \includegraphics[width=\linewidth]{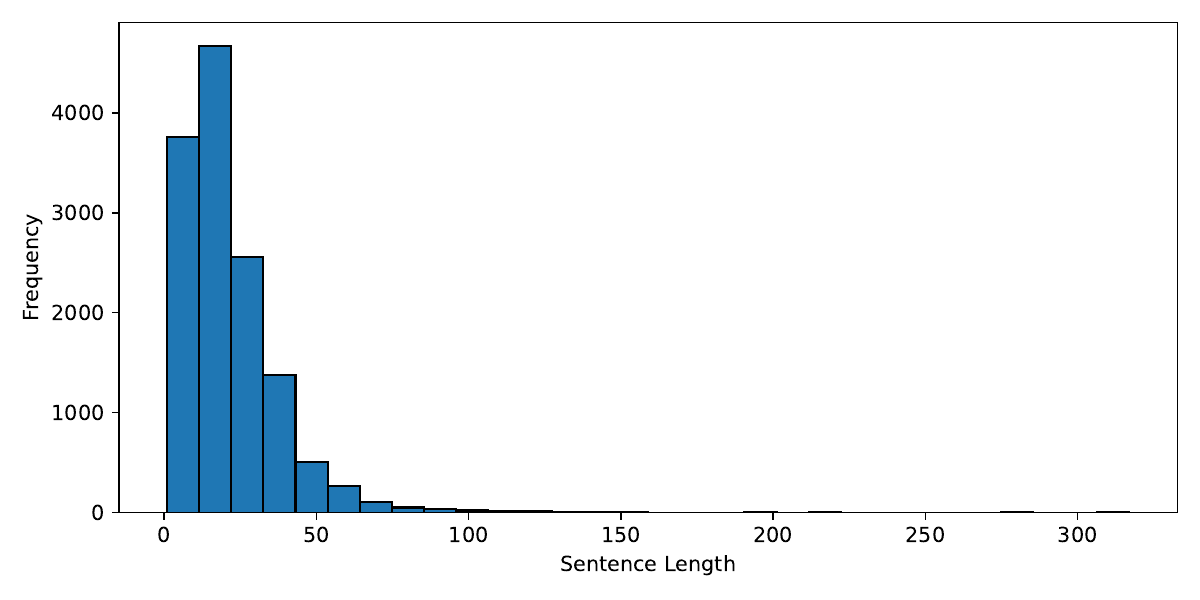}
    \caption{\texttt{French}}
  \end{subfigure}
  \vspace{0.3cm}
  \begin{subfigure}{0.48\linewidth}
    \includegraphics[width=\linewidth]{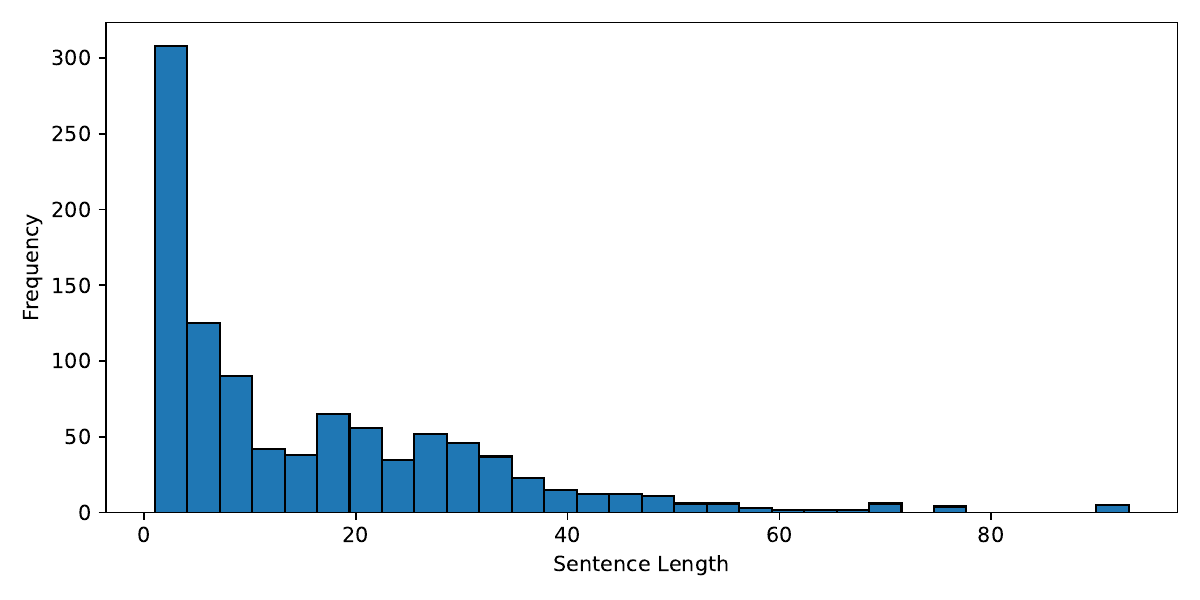}
    \caption{\texttt{Spanish}}
  \end{subfigure}
  \hfill
  \begin{subfigure}{0.48\linewidth}
    \includegraphics[width=\linewidth]{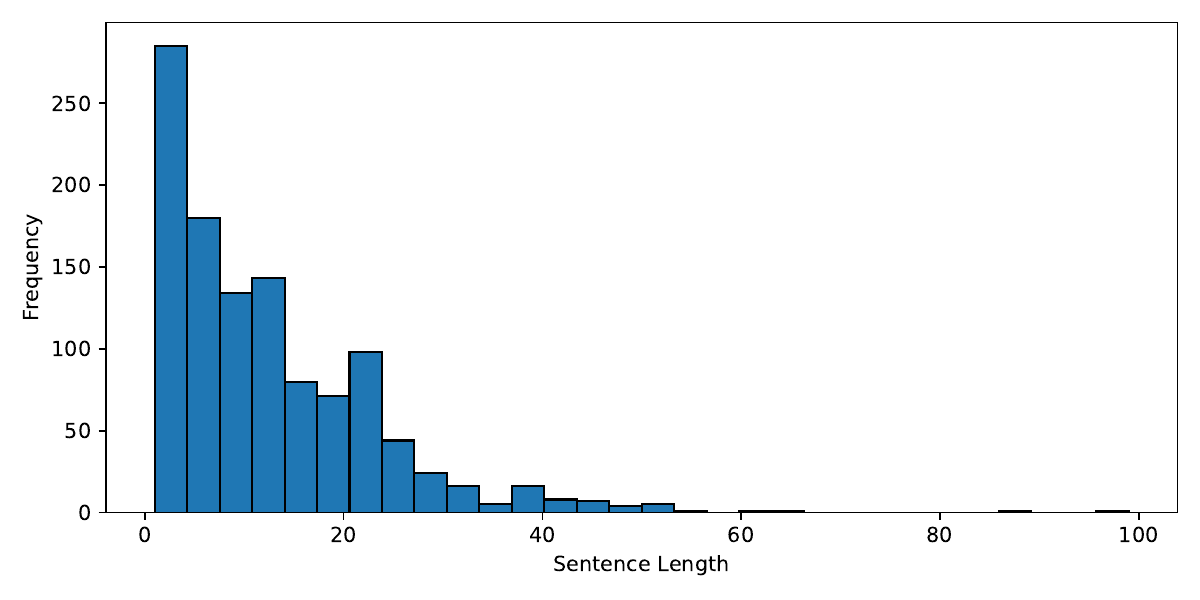}
    \caption{\texttt{Italian}}
  \end{subfigure}
  \caption{Sentence length distributions across languages: \texttt{Italian} and \texttt{English} exhibit longer sentences, consistent with higher redundancy and memorization potential.}
  \label{fig:sentence-lengths-hitz}
\end{figure}

\begin{figure}[t!]
  \centering
  \begin{subfigure}{0.48\linewidth}
    \includegraphics[width=\linewidth]{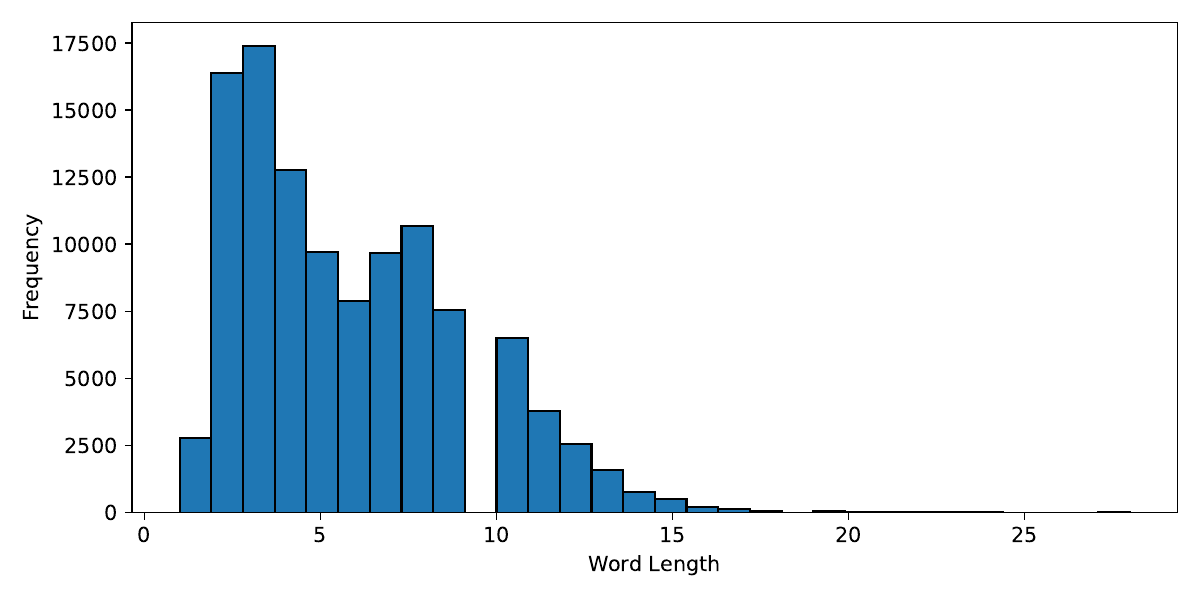}
    \caption{\texttt{English}}
  \end{subfigure}
  \hfill
  \begin{subfigure}{0.48\linewidth}
    \includegraphics[width=\linewidth]{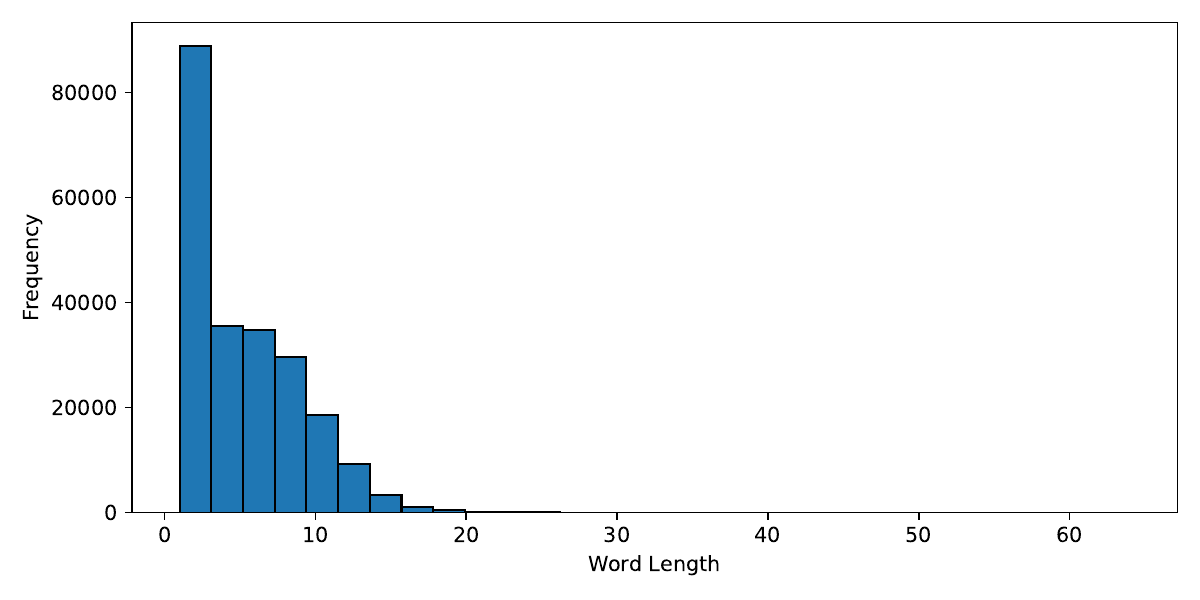}
    \caption{\texttt{French}}
  \end{subfigure}
  \vspace{0.3cm}
  \begin{subfigure}{0.48\linewidth}
    \includegraphics[width=\linewidth]{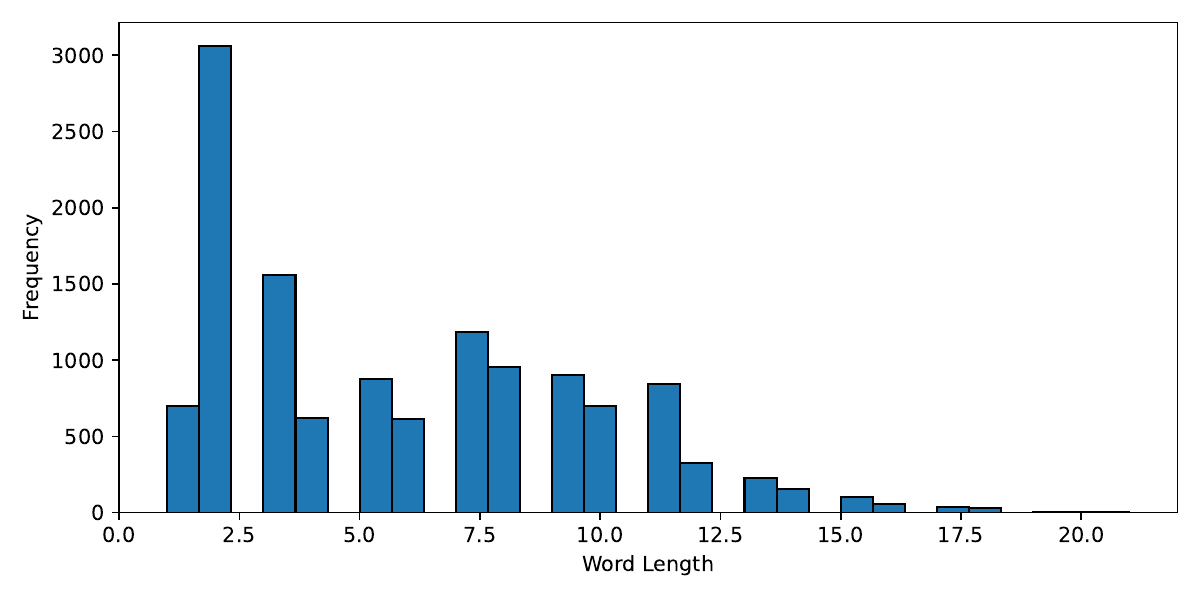}
    \caption{\texttt{Spanish}}
  \end{subfigure}
  \hfill
  \begin{subfigure}{0.48\linewidth}
    \includegraphics[width=\linewidth]{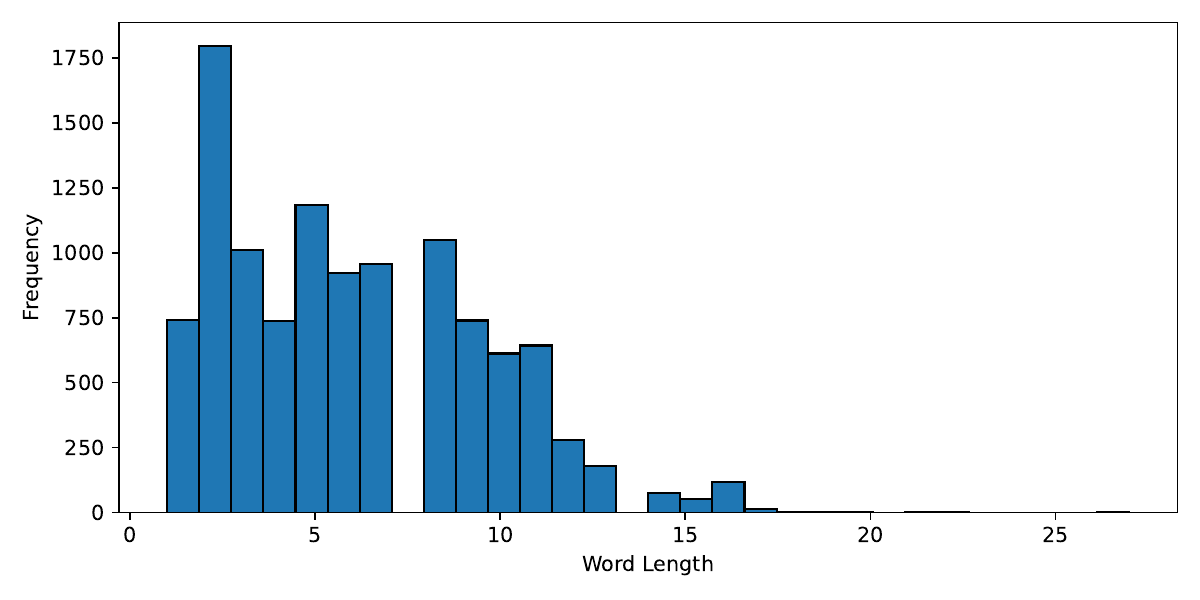}
    \caption{\texttt{Italian}}
  \end{subfigure}
  \caption{Word length distributions across languages: \texttt{Italian} and \texttt{Spanish} show heavier right tails, indicating longer lexical units and denser morphology.}
  \label{fig:word-lengths-hitz}
\end{figure}

To further characterize structural variability, we analyze sentence and word length distributions across languages (Figures~\ref{fig:sentence-lengths-hitz} and~\ref{fig:word-lengths-hitz}). Quantitatively, \texttt{Italian} exhibits the longest average sentence length ($\mu_{\text{sent}} \approx 23.4$ words), followed by \texttt{English} ($21.8$), \texttt{Spanish} ($20.7$), and \texttt{French} ($18.9$). This trend aligns with the higher redundancy and morphological density of Italian, suggesting broader contextual spans that may promote memorization. 

Similarly, word-length analysis reveals that \texttt{Italian} and \texttt{Spanish} have longer average words ($\mu_{\text{word}} = 5.99$ and $5.79$, respectively), while \texttt{French} ($5.48$) and \texttt{English} ($5.77$) remain slightly shorter and more evenly distributed. Italian’s longer words, coupled with its high redundancy (8.69), increase token-level repetition under subword tokenization, potentially heightening privacy risk. In contrast, French’s shorter sentences and lower capitalization rate ($7.3\%$) suggest a lower likelihood of memorizing personally identifiable terms or structured entities.

From a privacy standpoint, these quantitative differences highlight distinct trade-offs across languages. 
\texttt{Italian} shows the strongest exposure indicators: highest redundancy (8.69), longest sentences, and most extended word lengths, suggesting increased risk of memorization and entity leakage. 
\texttt{English} combines high syntactic entropy (2.90) and capitalization rate (14.5\%), which could heighten exposure to named entities and rare phrasing patterns. 
\texttt{Spanish}, while morphologically similar to Italian, demonstrates lower redundancy (7.38) and capitalization (9.7\%), implying moderate leakage susceptibility. 
\texttt{French} then exhibits the highest morphological complexity (1.34) but lowest vocabulary richness (0.078), favoring regular inflectional patterns that may mitigate verbatim recall.

\subsection{Linking Linguistic Characteristics to Privacy Vulnerabilities}
\label{sec:takeaways}

When contextualized with the corpus-level statistics from Table~\ref{tab:hitz_metrics} and the empirical findings in Sections~\ref{sec:extraction}–\ref{sec:mia}, a consistent picture emerges linking linguistic structure to privacy vulnerability. 
Leakage patterns observed across the three attack families: extraction, memorization, and membership inference, closely follow the typological properties of each language.

In the \textbf{extraction attack} (Section~\ref{sec:extraction}), both \texttt{Spanish} and \texttt{Italian} exhibit steady growth in leakage as prompt length increases. 
This behavior aligns with their higher \emph{redundancy} ($\mathcal{R}=7.38$ and $8.69$, respectively) and longer \emph{average word lengths} ($5.79$ and $5.99$), which encourage surface-level repetition and amplify memorization under subword tokenization. 
\texttt{English}, while less redundant, demonstrates pronounced leakage for short prompts, consistent with its high \emph{syntactic entropy} ($\mathcal{S}=2.90$), since even limited context can trigger memorized continuations. 
In contrast, \texttt{French}, with its greater \emph{morphological complexity} ($\mathcal{M}=1.35$) and shorter average sentences ($\mu_{\text{sent}} \approx 18.9$), displays a dampened extraction curve, suggesting that rich inflectional variability reduces exact sequence recall.

Results from the \textbf{counterfactual memorization} experiment (Section~\ref{sec:memorization}) reinforce these trends. 
\texttt{Italian} again shows the strong memorization signal, driven by its high redundancy and longer tokens that form stable phrase structures reused across contexts. 
\texttt{English} follows closely, where strong syntactic regularities facilitate verbatim recall of distinct patterns. 
\texttt{Spanish} exhibits low memorization; its structural diversity dilutes recurrence, while \texttt{French}, due to its inflectional diversity, maintains a relatively low recall rate for exact sequences, confirming that morphological variability provides natural resistance to overfitting.

Finally, under the \textbf{membership inference attack} (Section~\ref{sec:mia}), \texttt{English} fine-tunes show the clearest separation between “in” and “out” samples, indicating strong memorization and poor generalization. 
\texttt{Italian} also displays detectable separability, though less pronounced, whereas \texttt{Spanish} and especially \texttt{French} exhibit overlapping confidence distributions, reflecting smoother generalization and weaker membership signals. 
These trends parallel the corpus-level differences in redundancy and morphological diversity.

Overall, the quantitative correspondence between linguistic structure and empirical leakage across all attack types highlights that \textit{language itself is a determinant of privacy risk}. 
Languages with longer lexical units, higher redundancy, and predictable syntax (\texttt{English}, \texttt{Italian}) exhibit greater memorization and vulnerability to inference attacks. 
In contrast, morphologically rich and less redundant languages (\texttt{French}, \texttt{Spanish}) demonstrate improved privacy resilience, though longer prompts can still elevate extraction exposure. 



\section{Limitations}
\label{sec:limitations}

This study empirically examines how linguistic structure influences privacy leakage in LLMs, yet several limitations remain. 
Our experiments were conducted on relatively small multilingual medical corpora, which may limit generalizability; extending to larger datasets would improve robustness but requires substantial computational resources. 
The limited number of languages considered is also a limitation.
While we considered representative encoder and decoder architectures, exploring diverse model families and fine-tuning configurations could reveal further nuances. 
Finally, future work could assess fully multilingual models, rather than separately fine-tuned monolingual ones, to capture cross-lingual transfer effects, though this entails significant computational demands. 

\section{Conclusion}
\label{sec:conclusion}

We conduct a cross-linguistic analysis of privacy leakage in LLMs trained on distinct languages, showing that linguistic structure strongly influences model vulnerability. Across \texttt{English}, \texttt{Spanish}, \texttt{French}, and \texttt{Italian}, and under extraction, counterfactual memorization, and membership inference attacks, we observe clear structural effects: \texttt{Italian} shows the greatest leakage due to high redundancy and longer sentences, while \texttt{English} exhibits higher membership separability from greater syntactic entropy. In contrast, \texttt{French} and \texttt{Spanish} remain more resilient through richer morphology. These findings underscore the need for language-aware, structure-adaptive privacy defenses. 

\bibliographystyle{alpha}
\bibliography{sample}

\end{document}